%% file: acl_latex.tex
\pdfoutput=1

\documentclass[11pt]{article}

\usepackage[final]{acl}

\usepackage{times}
\usepackage{latexsym}

\usepackage[T1]{fontenc}

\usepackage[utf8]{inputenc}

\usepackage{microtype}

\usepackage{inconsolata}

\usepackage{multirow}
\usepackage{graphicx}
\usepackage{colortbl}
\usepackage{tabularx}
\usepackage{booktabs}
\usepackage{amsmath} 
\usepackage{microtype}
\usepackage{graphicx}
\usepackage{algorithm}
\usepackage{algpseudocode}
\usepackage{makecell}
\usepackage{xcolor}
\usepackage{enumitem}
\usepackage{hyperref}
\usepackage{tcolorbox}
\usepackage{longtable}
\usepackage{cleveref}
\usepackage[utf8]{inputenc}
\usepackage{multirow}
\usepackage{colortbl}
\usepackage{inconsolata}

\title{StraGo: Harnessing Strategic Guidance for Prompt Optimization}

\author{Yurong Wu\textsuperscript{1}\thanks{This work was done during an internship at Microsoft.}\quad Yan Gao\textsuperscript{2}\quad Bin Benjamin Zhu\textsuperscript{2}\quad Zineng Zhou\textsuperscript{3}\footnotemark[1] \quad Xiaodi Sun\textsuperscript{2}\\ 
\textbf{Sheng Yang\textsuperscript{1}\footnotemark[1]\quad Jian-Guang Lou\textsuperscript{2}\thanks{Corresponding author.}\quad Zhiming Ding\textsuperscript{1}\quad Linjun Yang\textsuperscript{2}}
\\
\textsuperscript{1}Institute of Software, CAS \& University of Chinese Academy of Sciences\\
\textsuperscript{2}Microsoft\\
\textsuperscript{3}Institute of Computing Technology, CAS \& University of Chinese Academy of Sciences\\
\{wuyurong20, zhouzineng22, yangsheng22\}@mails.ucas.ac.cn;\\ \{yan.gao, binzhu, sunstifler, jlou, Yang.Linjun\}@microsoft.com; zhiming@iscas.ac.cn\\
}

\begin{document}
\maketitle
\begin{abstract}
\input{parts/01_abstract}

\end{abstract}

\section{Introduction}
\input{parts/02_introduction}

\section{Methodology}
\input{parts/04_methodology}
\section{Experiments}

\input{parts/05_experiment}

\input{parts/03_related_work}

\input{parts/06_conclusion}

\bibliography{custom}

\appendix
\input{parts/appendix}

\end{document}

%% file: parts/01_abstract.tex
Prompt engineering is pivotal for harnessing the capabilities of large language models (LLMs) across diverse applications. While existing prompt optimization methods improve prompt effectiveness, they often lead to prompt drifting, wherein newly generated prompts can adversely impact previously successful cases while addressing failures. Furthermore, these methods tend to rely heavily on LLMs' intrinsic capabilities for prompt optimization tasks.
In this paper, we introduce \textsc{StraGo} (\textbf{Stra}tegic-Guided Optimization), a novel approach designed to mitigate prompt drifting by leveraging insights from both successful and failed cases to identify critical factors for achieving optimization objectives. \textsc{StraGo} employs a how-to-do methodology, integrating in-context learning to formulate specific, actionable strategies that provide detailed, step-by-step guidance for prompt optimization.
Extensive experiments conducted across a range of tasks, including reasoning, natural language understanding, domain-specific knowledge, and industrial applications, demonstrate \textsc{StraGo}'s superior performance. It establishes a new state-of-the-art in prompt optimization, showcasing its ability to deliver stable and effective prompt improvements.

%% file: parts/02_introduction.tex
Recent advancements in large language models (LLMs), such as ChatGPT and GPT-4, have significantly enhanced their analytical, reasoning, and contextual understanding capabilities~\cite{yue2024enabling,chang2023survey,xu2024survey}. LLMs are employed in various applications, such as Microsoft Copilot and New Bing, where users interact with the models through prompts. These prompts play a crucial role in guiding the LLMs' responses, ensuring outputs are accurate, relevant, and useful. However, the performance of LLMs heavily depends on prompt quality, and crafting effective prompts remains a complex, labor-intensive task that requires considerable expertise.

To overcome the challenge of crafting effective prompts, recent research has focused on creating and optimizing prompts automatically. Early approaches utilized reinforcement learning~\citep{deng2022rlprompt} or gradient-based methods~\citep{shin2020autoprompt}, though these techniques often require additional training or depend on the model’s internal state, limiting their applicability for API-based LLMs like ChatGPT and GPT-4. Recent studies have leveraged LLMs themselves as prompt generators~\cite{zhou2022large} or optimizers~\cite{yang2023large}. Advanced search algorithms, such as Monte Carlo Tree Search (MCTS)~\cite{wang2023promptagent} and evolutionary algorithms~\cite{guo2023connecting, fernando2023promptbreeder}, have also been applied to discover effective prompts. Additionally, some research has exploited the reflective capabilities of LLMs~\citep{shinn2023reflexion, chen2023teaching}, optimizing prompts by using erroneous examples to guide refinement, either explicitly or implicitly~\cite{pryzant2023automatic, yang2023large, hu2023evoke, tang2024unleashing}. These LLM-based optimization methods have demonstrated effectiveness across various tasks and hold promise for improving prompt quality.

However, search-based algorithms often suffer from inefficiency in prompt optimization due to the absence of a clear optimization direction at each step. Reflection-oriented methods aim to accelerate convergence by focusing on iteratively analyzing and correcting erroneous cases. However, concentrating on failure cases can sometimes negatively affect correct ones, especially when the errors exhibit outlier characteristics. Both search-based and reflection-oriented approaches can result in \emph{prompt drift}, where a newly generated prompt resolves certain failures but inadvertently disrupts previously successful cases.

Additionally, these methods typically provide the LLM with a task description and context without offering specific guidance on how to achieve the desired outcomes, relying solely on the LLM's inherent capabilities. For example, OPRO~\cite{yang2023large} supplies historical prompts with corresponding scores and task-specific data, expecting the LLM to generate more effective prompts. EvoPrompt~\cite{guo2023connecting} asks the LLM to merge two prompts into a new one without any instructions or strategy for doing so. Similarly, APO~\cite{pryzant2023automatic} presents erroneous cases and asks the LLM to correct them with new prompts, but without providing actionable guidance. This heavy reliance on the LLM's intrinsic abilities can be problematic for complex tasks, as the model may lack the necessary skills, leading to suboptimal prompt generation.

In this paper, we introduce \textsc{StraGo} (\textbf{Stra}tegic-Guided Optimization), a novel reflection-based prompt optimization method designed to overcome the limitations of existing approaches. Unlike prior methods, \textsc{StraGo} avoids bias towards failure cases by analyzing both successful and failed outcomes in each iteration, identifying key factors necessary for task success and understanding the causes of failures. Using this analysis, \textsc{StraGo} employs in-context learning to develop specific, actionable strategies that offer detailed, step-by-step guidance for prompt refinement. These strategies, combined with the analysis results, are used to optimize the prompt. Our extensive experiments across reasoning, natural language understanding, domain knowledge, and industrial applications demonstrate that this approach effectively corrects failures while minimizing adverse effects on successful cases. This unbiased iterative process, coupled with detailed guidance, achieves the best overall accuracy improvements post-optimization, setting a new state-of-the-art in prompt optimization.

Our major contributions are as follows:
\begin{enumerate}
    \item \textbf{Unbiased Reflective Optimization}: \textsc{StraGo} mitigates prompt drifting by incorporating both successful and failed cases in the optimization process, resulting in more stable and reliable prompt refinement.

    \item \textbf{Actionable Strategy Development}: \textsc{StraGo} leverages in-context learning to craft step-by-step, actionable strategies that guide prompt optimization, unlocking LLMs' potential in tasks where they initially lack sufficient expertise.

    \item \textbf{Broad Validation Across Diverse Tasks}: We extensively evaluate \textsc{StraGo} across various tasks, including reasoning, language understanding, domain-specific knowledge, and industrial applications, demonstrating that \textsc{StraGo} achieves state-of-the-art performance in prompt optimization.

\end{enumerate}

%% file: parts/04_methodology.tex
\subsection{Preliminaries}

\subsubsection{Task Formulation}

Given a task dataset $D$, our objective is to find the optimal prompt $p^*$ that enables an LLM to generate responses closely matching the desired outputs. This problem can be formalized as follows:
\begin{align}
\underset{p^*}{\min} \quad J(p^*) = \sum_{(x, y) \in D} \text{loss}(\text{LLM}(p^*, x), y),
\label{eq::task}
\end{align}
where $x$ and $y$ represent the input and its corresponding desired output from the task dataset $D$, and $p^*$ is the optimal prompt that minimizes the loss between the LLM's output and the desired output for all input-output pairs in $D$.

\subsubsection{Assessment Metrics}
\label{metrics}

Accuracy is the primary metric for evaluating the effectiveness of a prompt in solving a task using an LLM. However, during iterative prompt optimization, it is equally important to assess how new prompts affect both previously successful and failed cases. To capture this, we introduce two additional metrics: \emph{Adverse Correction Rate} (\textsc{Acr}) and \emph{Beneficial Correction Rate} (\textsc{Bcr}):

{\normalsize
\begin{align}
\textsc{Acr} &= \frac{\sum_{i=1}^{n} \mathbf{1} \big(p_{\text{pre}}(x_i) = y_i \wedge p_{\text{post}}(x_i) \neq y_i \big)}
{\sum_{i=1}^{n} \mathbf{1} \big(p_{\text{pre}}(x_i) = y_i \big)} \\
\textsc{Bcr} &= \frac{\sum_{i=1}^{n} \mathbf{1} \big(p_{\text{pre}}(x_i) \neq y_i \wedge p_{\text{post}}(x_i) = y_i \big)}
{\sum_{i=1}^{n} \mathbf{1} \big(p_{\text{pre}}(x_i) \neq y_i \big)}
\end{align}
}

where $p_{\text{pre}}(x_i)$ and $p_{\text{post}}(x_i)$ represent the model's predictions before and after optimization, respectively, for each input $x_i$ and its ground truth $y_i$. 

\textsc{Acr} measures the negative impact of optimization by capturing the proportion of correct predictions that become incorrect after applying the new prompt. In contrast, \textsc{Bcr} quantifies the positive impact by measuring the proportion of previously incorrect predictions that are corrected. Together with accuracy, these metrics offer a comprehensive evaluation of the new prompt's overall effectiveness, highlighting both its potential drawbacks and improvements.


\begin{figure*}
    \centering
    \includegraphics[width=\textwidth, trim=0cm 0cm 0cm 0cm]{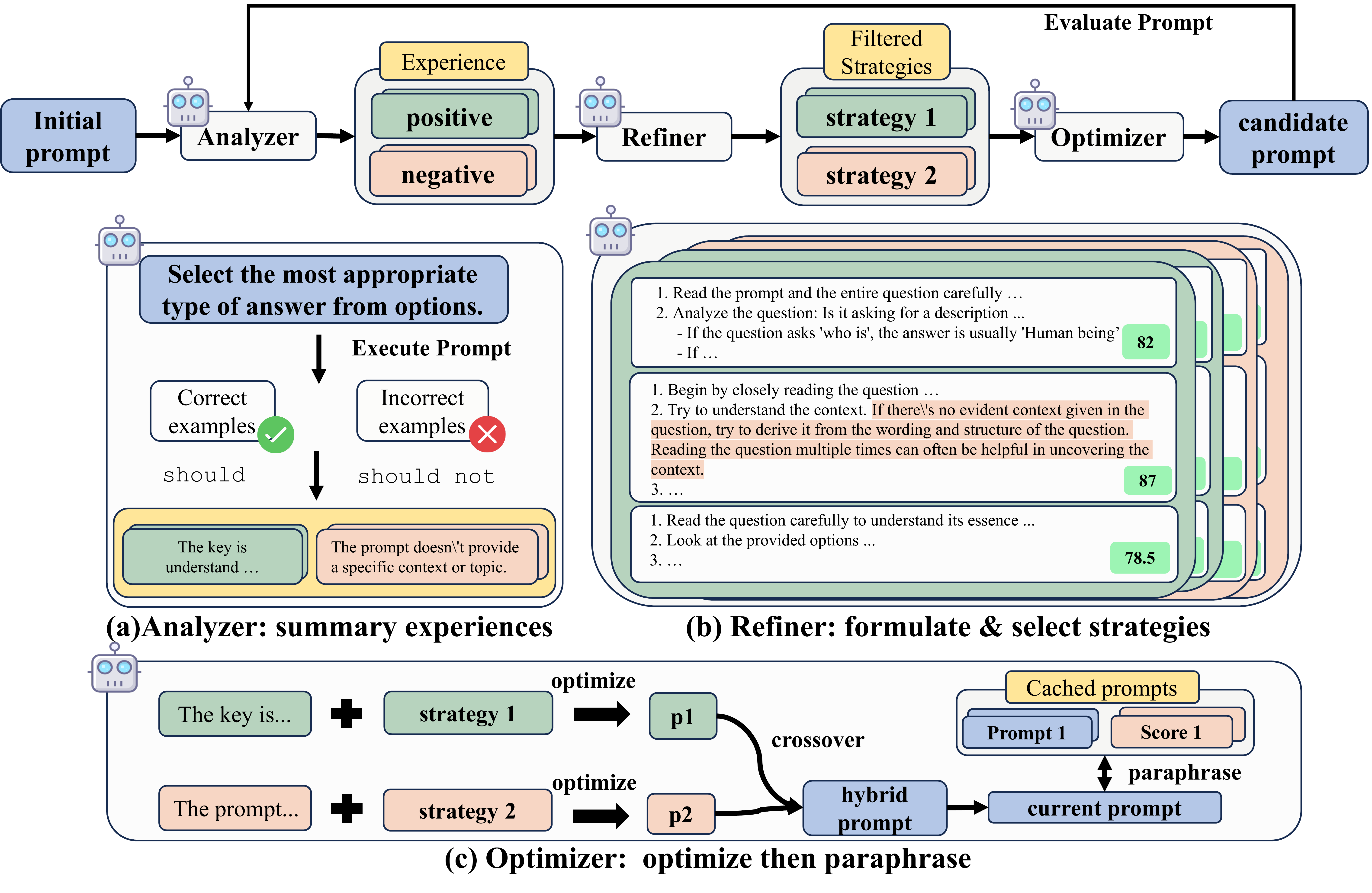}
    \caption{Flowchart of \textsc{StraGo}}
    \label{fig:framework}
\end{figure*}

\subsection{\textsc{StraGo}}


In each optimization iteration, \textsc{StraGo} samples both successful and failed cases to identify key factors for achieving task objectives and to understand why the current prompt leads the LLM to succeed or fail (Analyzer). Based on this analysis, it generates executable strategies that offer detailed, step-by-step guidance for optimization (Refiner). These strategies are then combined with the analysis results to optimize the prompt (Optimizer). Figure~\ref{fig:framework} illustrates the three main steps of \textsc{StraGo}, using the TREC task~\cite{voorhees2000building} as an example. Each module is discussed in detail in the following subsections. All meta prompts used in \textsc{StraGo} are provided in Appendix~\ref{app:meta prompts}.

\subsubsection{Analyzer}

\textsc{StraGo} differs from previous reflection-based methods by equally prioritizing the analysis of both correct and incorrect examples. Given a dataset $D = {(x_1,y_1), (x_2,y_2), \ldots, (x_n,y_n)}$, the model partitions it into two subsets after evaluation: $D_{correct}$ for correctly predicted samples and $D_{incorrect}$ for incorrectly predicted samples. From each subset, $K$ examples are selected for deep analysis. The Analyzer examines these selected examples to uncover the factors driving success in $D_{correct}$ and the reasons for failures in $D_{incorrect}$. These insights, termed \emph{positive experiences} and \emph{negative experiences}, guide LLMs by highlighting key actions to take and common errors to avoid. In our implementation, each example generates $M$ positive or negative experiences, depending on whether it belongs to $D_{correct}$ or $D_{incorrect}$.

\subsubsection{Refiner}





According to cognitive science principles~\cite{swanborn2010case}, humans typically approach problem-solving through three dimensions: identification (What it is), causation (Why it is), and method (How to do it). In this context, experiences relate to the identification dimension. While LLMs are generally capable of handling straightforward tasks, they may struggle with more complex challenges that require specific context or domain knowledge, as illustrated in Figure~\ref{fig:framework}(a), where the prompt lacks specific context or topic. To improve LLM performance in such cases, the Refiner adopts a two-step process: strategy formulation and strategy selection.

\textbf{Strategy formulation:} As noted by~\citet{ma2024large}, LLM-generated errors tend to follow specific patterns. For instance, miscalculations commonly occur in mathematical tasks, while misunderstandings or lack of contextual comprehension are frequent issues in language tasks. These patterns necessitate tailored strategies, making them ideal in-context learning demos. We focus on three prevalent error types: calculation errors in math tasks, misunderstandings in language tasks, and logical inference errors in reasoning tasks. We develop corresponding strategies for each error type and use them as in-context learning demos to help the LLM generate strategies that improve prompts based on both positive and negative experiences.

For each aforementioned error type, we select one or more representative examples. The LLM first generates an experience for each example and proposes a specific, actionable strategy to address it, which is then refined through manual revision. These examples, along with their associated experiences and strategies, serve as in-context learning demos, guiding the LLM in formulating detailed, step-by-step execution plans for both successful examples (positive experiences) and failed examples (negative experiences). In our implementation, we generate $N$ strategies for each example based on its experience. Figure~\ref{fig:framework}(b) illustrates three distinct strategies generated by the Refiner for $N=3$ to address a negative experience from a failed example.

\textbf{Strategy selection:} For the $N$ strategies generated by the Refiner for each example and its corresponding experience, we use an LLM to evaluate and score them based on criteria such as alignment, clarity, and feasibility. The strategy with the highest score is then selected to address the experience.

We assess the strategies across several dimensions: \textit{Match with Experience}, which evaluates how well the strategy addresses the identified issues; \textit{Clarity of Strategy}, which determines whether the strategy is clear and detailed; and \textit{Effectiveness in Addressing the Issue}, which measures the likelihood that the strategy will efficiently resolve the problem. To mitigate potential self-enhancement bias during evaluation~\citep{zheng2024judging}, we use a different LLM (Claude) for scoring. Additionally, following the scoring method used in~\citet{thomas2023large}, we conduct five  assessments with the LLM and average their scores for enhanced stability and reliability. Figure~\ref{fig:framework}(b) presents the averaged score for each of the three strategies addressing a negative experience from a failed example, with the highest-scoring strategy (shown in the middle) being selected.

\subsection{Optimizer}

Although LLMs can process long text inputs, they often struggle to thoroughly consider every detail when handling both positive and negative experiences, along with their associated strategies. To mitigate this issue, we implement an optimization method that processes these experiences separately and then combines them through a crossover procedure. The optimizer operates in three main steps: Optimize, Crossover, and Paraphrase.

\textbf{Optimize:} For each selected successful or failed example, the Analyzer generates $M$ positive or negative experiences. The Refiner then generates a strategy for each experience, and the Optimizer creates a revised prompt based on the strategy. These revised prompts are divided into two sets: one for prompts derived from positive experiences and the other for those based on negative experiences.

\textbf{Crossover:} Following the approach of~\citet{guo2023connecting}, which shows that combining LLMs with evolutionary algorithms can improve prompt fusion (similar to genetic algorithms), we select two prompts, one from each set, and perform a crossover operation to produce a hybrid prompt.

\textbf{Paraphrase:} A cache is maintained to store the top \textit{n} prompts and their corresponding scores from previous evaluations on a validation set. Each hybrid prompt is paraphrased using the prompts in the cache, and both the paraphrased and hybrid prompts are evaluated as candidate prompts. The best prompt is either selected for the next iteration of optimization if the stopping condition has not been met or output as the optimized prompt. The cache is then updated with the evaluation results.

%% file: parts/05_experiment.tex
\subsection{Evaluation Tasks}

We select five relatively challenging tasks from BBH~\cite{suzgun2022challenging}, chosen for their historically low performance scores, though they are among the simpler tasks included in our evaluation.

In addition to these tasks, we incorporate two well-known natural language understanding (NLU) tasks: SST-5~\cite{socher2013recursive}, a sentiment classification task based on movie reviews, and TREC~\cite{voorhees2000building}, which identifies types of responses. We also include MedQA~\cite{jin2021disease} and MedMCQA~\cite{pal2022medmcqa} to evaluate our method's effectiveness in tasks related to medical and pharmacological knowledge.

To evaluate the effectiveness of our method in industrial scenarios, we select an internal personalized search task named \emph{Personalized Intent Query}. This task uses anonymized search data to determine whether non-personalized search results should be reordered based on user-specific information such as location, language, and search history. The task involves step-by-step initial prompts typical of complex industrial tasks and diverse, extensive data content that often includes redundant information. These characteristics represent common challenges in industrial-level prompt optimization. For detailed data specifications, please refer to Appendix~\ref{appendix:data}.

\input{tables/main_result}

\subsection{Baselines}
The following prompt optimization methods serve as baselines for comparison with our method:
\begin{itemize}
    \item \textbf{CoT}: CoT~\cite{wei2022chain, kojima2022large} is a popular baseline in many studied. In our setup, CoT is initiated by appending the phrase "Let's think step by step." after the question without utilizing any examples.
    \item \textbf{APO}: APO~\cite{pryzant2023automatic} generates natural language-level gradients from incorrect examples and uses these gradients to reverse-edit the prompt. APO represents explicit feedback methods.
    \item \textbf{OPRO}: OPRO~\cite{yang2023large} utilizes implicit feedback by tracking a historical trajectory of previous prompts and their associated scores. During prompt optimization, OPRO leverages these trajectories to guide the LLM in generating prompts aimed at achieving higher scores.
    \item \textbf{EvoPrompt}: EvoPrompt~\cite{guo2023connecting} applies evolutionary algorithms, such as genetic algorithms and differential evolution, to generate prompts that optimize performance on validation sets. It serves as a representative method for search-based optimization techniques.
\end{itemize}

\subsection{Experimental Details}

We conduct extensive experiments using GPT-4~\cite{achiam2023gpt} to evaluate the effectiveness of \textsc{StraGo} and the baseline methods. APO, OPRO, and \textsc{StraGo} all start with the same initial prompt, while EvoPrompt uses 14 additional variations.

A subset of the test set is selected as the validation set for prompt optimization. In each iteration, the validation set is used to assess prompt quality. During the final testing phase, the remaining test samples are used to evaluate the optimized prompts. For each method, we select the top 5 optimized prompts with the highest validation scores and evaluate them on the test samples, reporting the performance of the best-performing prompt. For \textsc{StraGo}, we set $K$, $M$ and $N$ to 3. To ensure consistent evaluation, the temperature is set to 0. As outlined by~\citet{ma2024large}, all methods perform approximately the same number of prompt searches. Detailed parameter settings are provided in Appendix~\ref{app:setting}.

\subsection{Main Results}

The experimental results are reported in Table~\ref{tb:main_result}, where \textsc{StraGo} consistently outperforms all baselines across the six tasks, showcasing the effectiveness of our approach.

\paragraph{Performance on BBH and NLU tasks.}
\textsc{StraGo} achieves 79.77\% accuracy on BBH, 56.34\% on SST-5, and 87.21\% on TREC, surpassing previous state-of-the-art (SOTA) methods by 2.37\%, 0.82\%, and 2.31\%, respectively. These results demonstrate \textsc{StraGo}'s strong performance on relatively straightforward tasks. In contrast, EvoPrompt shows smaller improvements than APO and OPRO on BBH and TREC, suggesting that search-based methods like EvoPrompt may face challenges in rapid convergence. This highlights the importance of precise and targeted optimization strategies for rapid convergence in iterative prompt optimization.

\paragraph{Performance on Domain-specific Tasks.}
A notable trend is that on domain-specific tasks like MedQA and MedMCQA, all baselines show limited improvements, with none exceeding 1\%. Some methods, particularly EvoPrompt, even exhibit performance declines, likely because they don't leverage feedback from the data. In domain-specific tasks, relying solely on LLMs' intrinsic capabilities often fails to yield prompts well-suited to the data's characteristics. In contrast, \textsc{StraGo} demonstrates improvement, with a 1.22\% gain on MedQA and a 1.33\% gain on MedMCQA. This suggests that \textsc{StraGo}'s step-by-step prompt-revising strategy is more effective at inducing relevant domain knowledge and generating prompts tailored to the specific expertise required for these tasks.

\input{tables/analysis}

\paragraph{Performance on Industrial Scenario Tasks.}
In the Personalized Intent Query task, we compare \textsc{StraGo} only with APO due to the unique characteristics of its data. As shown in Table~\ref{tb:main_result}, APO experiences performance degradation when processing step-by-step instructions, likely because it struggles to accurately identify the specific steps that require editing in lengthy directives. In contrast, \textsc{StraGo} achieves a 2.16\% performance improvement, demonstrating that its approach of incrementally integrating experiences while formulating step-by-step strategies provides valuable contextual information for optimization. 

In summary, \textsc{StraGo} proves effective not only for simple prompts but also for addressing complex tasks, including those encountered in industrial scenarios.

\section{Analysis}

\subsection{Data Analysis}

To validate the importance of maintaining correctly predicted samples while correcting mispredicted ones during prompt optimization, we analyze the prompt drifting effect of each optimization method. Specifically, we compare the final prompts generated by various methods with the initial prompts, assessing how many new errors an optimized prompt introduces while correcting existing ones. The results are reported in Table~\ref{tb:analysis}\footnote{Note that the denominators for calculating \textsc{Acr} and \textsc{Bcr} differ.}

As shown in Table~\ref{tb:analysis}, \textsc{StraGo} exhibits the lowest \textsc{Acr} and the highest \textsc{Bcr} for four of the six tasks, indicating that its optimized prompts correct more erroneous samples while adversely affecting fewer correctly predicted samples than the baseline methods. This demonstrates \textsc{StraGo}'s superior performance compared to the baselines. The impact of maintaining correct samples is particularly significant in tasks with high-quality initial prompts. For instance, in MedQA, where the accuracy is 77.83\%, although APO corrects more errors than \textsc{StraGo} (34.62\% or 90 erroneous samples compared to 26.92\% or 70 erroneous samples), it also adversely affects more correct samples (10.41\% or 95 correct samples compared to 4.49\% or 41 correct samples). This results in a decline in performance for APO compared to \textsc{StraGo}, as shown in Table~\ref{tb:main_result}. We attribute this to \textsc{StraGo}'s integration of correct examples and positive experiences during prompt optimization, which helps avoid significant deviations from overall task objectives, especially when the initial prompt is already effective.

\subsection{Ablation Study}

We conduct an in-depth analysis on two tasks: the readily optimized TREC task and the domain knowledge-intensive MedMCQA task. In this study, we systematically remove both positive and negative experiences from the Analyzer, as well as strategies from the Refiner. The experimental results are presented in Table~\ref{tb:abalation}.

\paragraph{The Impact of Experience.} The results in Table~\ref{tb:abalation} indicate that removing positive experiences significantly increases \textsc{Acr}, leading to performance declines for \textsc{StraGo} across both tasks. This underscores the critical role of positive experiences in maintaining correctly predicted samples and enhancing overall task performance. Additionally, comparisons with Table~\ref{tb:main_result} reveal that \textsc{StraGo}, when utilizing only positive experiences and strategies, can effectively optimize performance, consistently outperforming all baseline methods in the TREC task. Conversely, eliminating negative experiences results in a reduction in \textsc{Bcr}, indicating that these experiences provide vital information for correcting erroneous samples and adapting to the subset of mispredicted data. Their absence impairs the Optimizer's ability to modify the prompt effectively, hindering the incorporation of pivotal text relevant to this data subset.

\paragraph{The Impact of Strategies.} Analysis of Table~\ref{tb:abalation} reveals that \textsc{StraGo} maintains robust performance in simpler tasks even without explicit strategies. However, the omission of strategies significantly diminishes \textsc{StraGo}'s effectiveness in tasks requiring domain knowledge. This disparity may arise from the fact that, in simpler tasks, LLMs can leverage their inherent capabilities to extract useful knowledge for prompt optimization. In contrast, these capabilities are often insufficient for knowledge-intensive tasks. By integrating explicit execution strategies, \textsc{StraGo} enhances the LLMs’ ability to engage in deeper analytical thinking, uncovering more domain-specific insights and providing valuable guidance for the Optimizer.

\input{tables/ablation}

\subsection{Convergence Analysis}
\begin{figure}
    \centering
    \includegraphics[width = \linewidth, trim=0cm 0cm 0cm 0cm, clip=true]{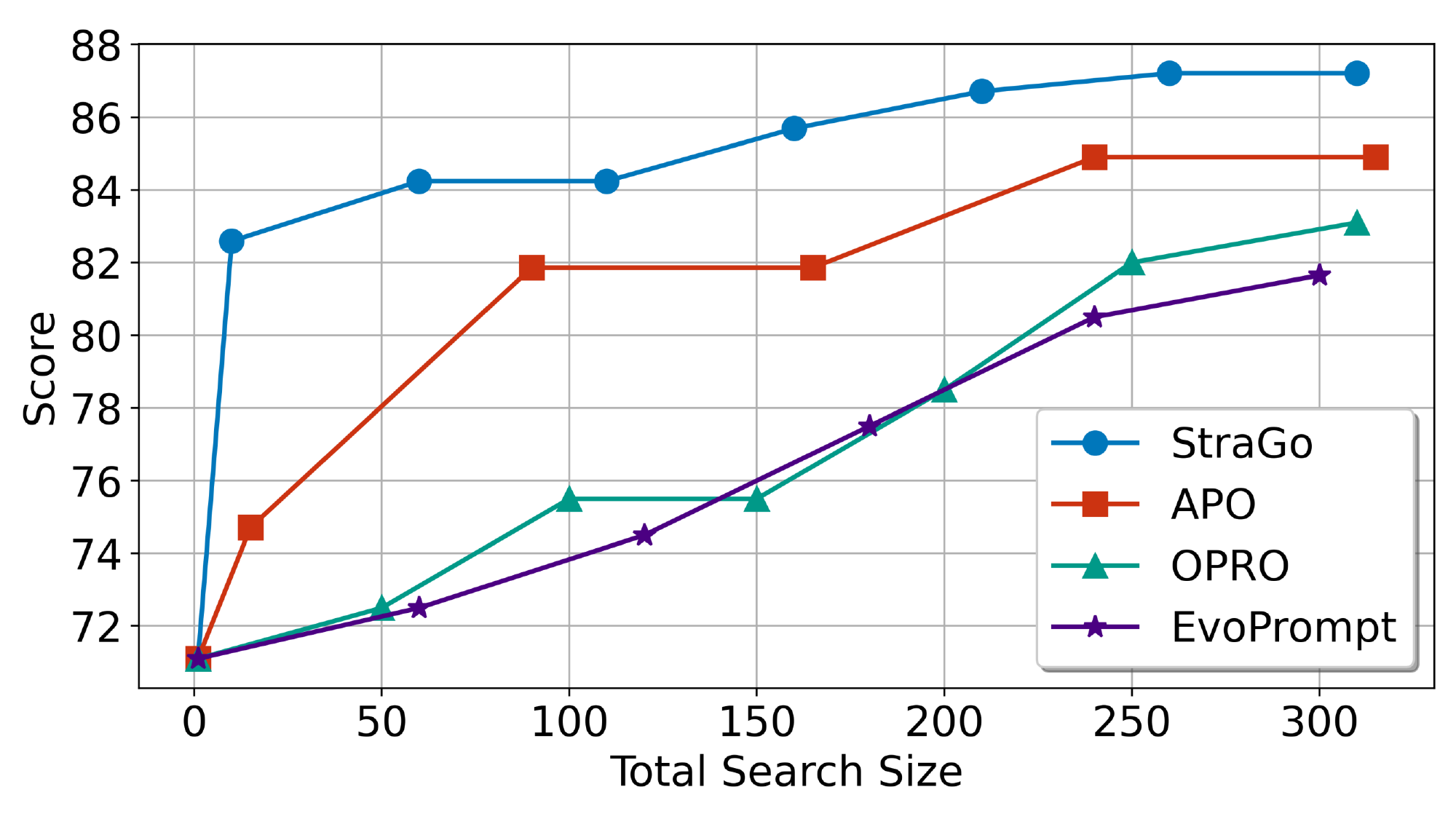}
    \caption{Convergence curves for the TREC task: Comparison of test set scores for the optimal prompt across different search sizes and various prompt optimization methods.}
    \label{fig:convergence}
\end{figure}

We analyze the convergence of \textsc{StraGo} in comparison to the three baseline methods on the TREC task, with results shown in Figure~\ref{fig:convergence}. Notably, \textsc{StraGo} converges significantly faster than the baseline methods. For example, to achieve a test set score above 80\%, \textsc{StraGo} requires the exploration of only 10 prompts, whereas methods like APO need over 90 prompts. This rapid convergence is likely due to \textsc{StraGo} providing more valuable reference information than its counterparts. In a single optimization cycle, the Optimizer not only utilizes positive and negative experiences but also incorporates corresponding strategies. This approach allows the Optimizer to access more comprehensive information and generate prompts with enhanced generalization capabilities.

\subsection{Cost Analysis}


We compare the resource consumption of \textsc{StraGo} with that of baseline methods by estimating the number of API calls and total token usage (see Appendix~\ref{app:cost} for detailed estimation methods). The results for the TREC dataset are presented in Table~\ref{tb:cost}. As shown, APO requires the fewest API calls, followed closely by \textsc{StraGo}. Unlike OPRO and EvoPrompt, both of these methods leverage the UCBandit algorithm to filter out many candidate prompts, thus reducing evaluation costs on the validation set. In terms of token consumption, EvoPrompt and \textsc{StraGo} exhibit the highest usage. EvoPrompt's elevated consumption arises from the need to evaluate numerous candidate prompts on the validation set, while \textsc{StraGo}'s higher usage is due to the longer length of its optimized prompts compared to other methods. However, given \textsc{StraGo}'s significant performance improvement (from 84.90\% to 87.21\%) over the other methods, this resource expenditure is considered justified.

\input{tables/cost}

\input{tables/ex2stra}

\subsection{Performance Using Different Models}

We evaluate \textsc{StraGo} and baseline methods using GPT-3.5-turbo and GPT-4 as evaluators (the task models used to assess prompt performance), with GPT-4 also serving as the optimizer (the model used to enhance the prompt). The experimental results are reported in Table~\ref{tb:model}. \textsc{StraGo} achieves a performance improvement of 5.45\% on GPT-3.5-turbo and 10.34\% on GPT-4. Additionally, \textsc{StraGo} outperforms the best baseline by 1.86\% on GPT-3.5-turbo and 2.37\% on GPT-4. This suggests that \textsc{StraGo} performs better with more advanced models. The superior performance observed with GPT-4 may be attributed to its improved adherence to instructions compared to GPT-3.5-turbo, which appears to struggle with capturing specific instructional nuances, even with a finely tuned strategic prompt. This phenomenon has been noted in other studies~\cite{zeng2023evaluating, ma2024large}. Detailed results can be found in Appendix~\ref{app:detail_results}.

\input{tables/model}

\section{Case Study}

This section provides an in-depth examination of the strategies developed by the Refiner and the optimization processes undertaken by the Optimizer, as illustrated through tow cases detailed in Table~\ref{tb:ex2stra}.

The first case pertains to a movie recommendation task. In this scenario, the Analyzer identifies the prompt’s failure, attributing it to the absence of a clear similarity criterion. To rectify this, the Refiner develops a strategy focusing on identifying such criteria, particularly by scrutinizing outlier data. Subsequently, the Optimizer refines the prompt by addressing this diagnosed error and integrating the strategic insights.

The second case involves the Snarks task, where the Analyzer underscores the importance of focusing on contextual clues, such as specific words or phrases. The Refiner then crafts a strategy that not only incorporates these basic experiential insights but also emphasizes the analysis of sentence tone, specifically to discern exaggeration or overstatement. These additional insights are pivotal in determining the ironic intent of a sentence.

%% file: tables/main_result.tex
\begin{table*}[ht]
    \centering
    \setlength\tabcolsep{4pt} 
    \renewcommand{\arraystretch}{1.2} 
    \scalebox{0.9}{ 
    \begin{tabular}{ccccccc}
        \toprule
        \textbf{Method} & \textbf{BBH} & \textbf{SST-5} & \textbf{TREC} & \textbf{MedQA} & \textbf{MedMCQA} & \textbf{Per. Query}\\
        \midrule
        MI (Manual Instructions)  & - & 54.48 & 71.10 & 77.83 & 65.87 & 67.97 \\
        CoT \cite{wei2022chain} & 69.43 & 53.86 & 64.40 & 49.10 & 59.07 & -\\
        APO \cite{pryzant2023automatic}  & 76.50 & 55.52 & 84.90 & 77.41  & 65.93 & 67.10 \\
        OPRO \cite{yang2023large} & 77.40 & 55.31 & 83.10 & 76.56 & 66.00 & - \\
        EvoPrompt \cite{guo2023connecting} & 75.48 & 55.15 & 81.65 & 77.15 & 65.47 & -\\
        \textsc{StraGo} (Ours) & \textbf{79.77} & \textbf{56.34} & \textbf{87.21} & \textbf{80.05}& \textbf{67.20} & \textbf{69.26} \\
        \bottomrule
    \end{tabular}}
    \caption{Performance across six tasks using GPT-4 as the evaluator with Q\_END zero-shot evaluation results. The initial instruction is CoT for BBH and the manual instructions for the other tasks. Bold text indicates the best performance achieved.}
    \label{tb:main_result}
\end{table*}

%% file: tables/analysis.tex

\begin{table*}[!t]
    \centering
    \renewcommand{\arraystretch}{1.2}  
    \setlength{\tabcolsep}{5pt}
    \begin{tabular}{ccccccccccccc}
        \toprule
        & \multicolumn{2}{c}{\textbf{BBH}} & \multicolumn{2}{c}{\textbf{SST-5}} & \multicolumn{2}{c}{\textbf{TREC}} & \multicolumn{2}{c}{\textbf{MedQA}} & \multicolumn{2}{c}{\textbf{MedMCQA}} & \multicolumn{2}{c}{\textbf{Per. Query}}\\
        \cmidrule{2-13}
        \multicolumn{1}{c}{\multirow{-2}{*}{\textbf{Method}}} & \textsc{Acr} & \textsc{Bcr} & \textsc{Acr} & \textsc{Bcr} & \textsc{Acr} & \textsc{Bcr} & \textsc{Acr} & \textsc{Bcr} & \textsc{Acr} & \textsc{Bcr} & \textsc{Acr} & \textsc{Bcr} \\
        \midrule
        APO       & 7.77 & 40.57 & 8.48 & 12.42 & \underline{3.56} & 56.5 & 10.41 & \textbf{34.62} & 5.16 & 9.96 & 6.37 & 10.81\\
        OPRO      & 7.53 & 41.59 & 9.11 & 12.73 & 4.57 & 53.75 & 8.87 & 25.38 & 5.36 & 10.74 & - & -\\
        EvoPrompt & 8.72 & 37.29 & 8.10 & 11.21 & 5.08 & 50.00 & 9.30 & 24.61 & 4.66 & 7.81 & - & -\\
        \textsc{StraGo} & \underline{4.59} & \textbf{49.98} & \underline{7.47} & \textbf{13.03} & 3.86 & \textbf{65.25} & \underline{4.49} & 26.92 & \underline{4.35} & \textbf{12.3} & \underline{3.82} & \textbf{12.12}\\
        \bottomrule
    \end{tabular}
    \caption{\textsc{Acr} and \textsc{Bcr} values for different optimization methods. The results for BBH represent the average across the five subtasks. Underlined values indicate the smallest (best) \textsc{Acr}, while bold values denote the largest (best) \textsc{Bcr}.}
    \label{tb:analysis}

\end{table*}

%% file: tables/ablation.tex
\begin{table}[!t]
    \centering
    \renewcommand{\arraystretch}{1}  
    \resizebox{\linewidth}{!}{
    \begin{tabular}{clccc}
        \toprule
        \textbf{Task} & \textbf{Method} & \textsc{Acr} & \textsc{Bcr} & \textsc{Acc.} \\
        \midrule
        \multirow{4}{*}{TREC} & Ours  & 3.86 & 65.25 & 87.21 \\ 
                            &\hspace{2pt}\textit{w/o.} pos.    & 4.67 & 56.75 & 84.18   \\
                            &\hspace{2pt}\textit{w/o.} neg.  & 4.17 &61.5 & 85.62     \\
                            &\hspace{2pt}\textit{w/o.} strat.    & 4.27 & 59.25 & 85.19 \\
        \midrule
        \multirow{4}{*}{MedMCQA} & Ours  & 4.35 &  12.3 & 67.20 \\ 
                            &\hspace{2pt}\textit{w/o.} pos.       & 8.10 & 16.02 & 66.00 \\
                            &\hspace{2pt}\textit{w/o.} neg.       & 4.15 & 9.96 & 66.53 \\
                            &\hspace{2pt}\textit{w/o.} strat.      & 5.06 & 9.18 & 65.67 \\

        \bottomrule
    \end{tabular}}
    \caption{Results of the ablation study on TREC and MedMCQA tasks: Impact of omitting positive experiences (w/o pos.), negative experiences (w/o neg.), and all strategies (w/o strat.) on \textsc{StraGo}.}
    \label{tb:abalation}
\end{table}

%% file: tables/cost.tex
\begin{table}[ht]
    \centering
    \setlength\tabcolsep{2pt} 
    \renewcommand{\arraystretch}{1} 
    \begin{tabular}{ccccc}
        \toprule
         & \textbf{API} & \textbf{Tokens} & \textbf{Search Size} & \textbf{Score}\\
        \midrule
        APO & 18.4K & 3.31M & 315 & 84.90\\
        OPRO & 93.3K & 4.11M & 310 & 83.10\\
        EvoPrompt & 61.3K & 5.46M & 300 & 81.65\\
        \textsc{StraGo}  & 23.8K & 5.46M & 310 & 87.21 \\
        \bottomrule
    \end{tabular}
    \caption{Cost comparison on TREC Task.}
    \label{tb:cost}
\end{table}

%% file: tables/ex2stra.tex
\definecolor{green}{RGB}{100,210,190}

\begin{table*}[!t]
\centering
\scalebox{0.72}{
\begin{tabular}{lp{1.1\textwidth}}
\toprule
\textbf{Task}                & \textbf{Experience and Corresponding Strategy}  \\\midrule
\multirow{10}{*}{\makecell{Movie \\ Recommendation}} & \textbf{Negative Experience:} Ambiguity in the similarity criterion: The prompt asks to find a movie similar to a given set of movies without specifying the basis of similarity ...  \\\cmidrule{2-2}
 & \textbf{Strategy:} ... 3. Recognize Outliers: \textcolor{red}{Also pay attention to the odd ones out, or the movies that don't share the above listed common factors. This could potentially give hints on what the 'similarity' criterion could be} ... \\ \cmidrule{2-2}
 & \textbf{Refined Prompt:} Let's carefully analyze each step. Given a set of movies, \textcolor{green}{determine which option is most similar based on common factors such as genre, era, theme, actors, or director. Be aware of any outliers in the set, as this might give hints on what the 'similarity' criterion could be}. Remember, in the absence of clear instructions or when there are multiple potential correct answers, it might be necessary to make an educated guess. Let's begin. \\ \midrule

\multirow{8}{*}{Snarks}  & \textbf{Positive Experience:} Contextual Clues: Notice words or phrases that may indicate sarcasm. This can be a circumstance or expectation that sounds out of ordinary, such as losing money for winning in example.\\\cmidrule{2-2}
 & \textbf{Strategy:} ... 2. \textcolor{red}{Look for any words or phrases that contradict usual or expected situations} ... 3. \textcolor{red}{Pay special attention to the tone of the statement} ... 4. \textcolor{red}{Locate any exaggeration or hyperbole in the statement} ...  \\\cmidrule{2-2}
 & \textbf{Refined Prompt:} Please identify the sarcastic statement from the given options. \textcolor{green}{Remember, sarcasm often involves statements that contradict usual situations or expectations and has a mocking or scornful tone. Look for statements that are implausible or absurd under normal circumstances and note any exaggerations or hyperbole.} The context of the statement can also help you understand its sarcastic nature. \\\bottomrule  
\end{tabular}
}
\caption{Two cases illustrating the strategy and optimization of \textsc{StraGo}. Note that the refined prompts displayed do not represent the best optimization result.}
\label{table:case}
\label{tb:ex2stra}
\end{table*}

%% file: tables/model.tex
\begin{table}[ht]
    \centering
    \setlength\tabcolsep{4pt} 
    \renewcommand{\arraystretch}{1} 
    \begin{tabular}{ccc}
        \toprule
         & \textbf{GPT-3.5-turbo} & \textbf{GPT-4} \\
        \midrule
        CoT & 56.37 & 69.43 \\
        APO & 59.96 & 76.50 \\
        OPRO & 59.78 & 77.40 \\
        EvoPrompt & 59.67 & 75.48 \\
        \textsc{StraGo}  & 61.82 & 79.77 \\
        \bottomrule
    \end{tabular}
    \caption{Performance of GPT-3.5-turbo and GPT-4 on BBH Task.}
    \label{tb:model}
\end{table}

%% file: parts/03_related_work.tex
\section{Related Work}
\label{sect::related_work}

\subsection{Automatic Prompt Engineering}

Prompt optimization aims to discover the most effective prompts for specific tasks~\cite{sahoo2024systematic, liu2023pre}. Initially, this optimization relied heavily on manually crafted templates designed by experts~\cite{white2023prompt}, which is labor-intensive, especially for complex tasks. To address this, researchers have developed various automated optimization techniques, broadly categorized into discrete and continuous methods~\cite{li2021prefix, zhang2021differentiable}. Discrete optimization modifies the prompt text by adjusting specific tokens. For instance, a prompt like "Let's think step by step" could be modified to "Take a deep breath and work on this problem step-by-step"~\cite{yang2023large}. Continuous optimization, in contrast, manipulates prompt embeddings by appending a latent space vector to the start of the embedding~\cite{lester2021power, Wen_Jain_Kirchenbauer_Goldblum_Geiping_Goldstein_2023}. Our approach, \textsc{StraGo}, focuses on editing discrete text without requiring additional training.

\subsection{LLM-based Prompt Optimization}

Recent studies increasingly utilize LLMs for prompt optimization~\cite{zhou2022large}. Advanced search techniques, such as Monte Carlo Tree Search (MCTS)~\cite{wang2023promptagent} and evolutionary algorithms~\cite{guo2023connecting, fernando2023promptbreeder}, are employed to iteratively refine and integrate potential candidate prompts, enhancing their effectiveness. Additionally, some research leverages the self-reflective capabilities of LLMs, generating prompts that preemptively avoid errors by analyzing incorrect examples and their underlying causes~\cite{pryzant2023automatic, yang2023large, ye2023prompt, tang2024unleashing}. This reflective approach allows models to learn from past mistakes, improving both the accuracy and relevance of future prompts.

%% file: parts/06_conclusion.tex
\section{Conclusion}

In this paper, we introduce \textsc{StraGo}, a strategy-guided, reflective-based optimization method that utilizes balanced iterations to analyze both successful and failed cases. This innovative approach identifies critical factors for achieving objectives while providing insights into the reasons for failures. By leveraging in-context learning, \textsc{StraGo} delivers detailed, step-by-step guidance for prompt optimization. Experiments conducted across diverse tasks—ranging from simple scenarios to domain-specific and complex industrial contexts—demonstrate that \textsc{StraGo} significantly outperforms existing prompt optimization methods, establishing a new state-of-the-art in the field.

\section{Limitations}

Our limitations are outlined as follows:


\paragraph{Fairness of Comparison:} To ensure fair comparisons, we adjust certain parameters in the official code of baseline methods, aligning the number of searches across all methods to approximately 300-315. While slight variations in the number of searches may exist between methods, these differences are minimal and within an acceptable range to maintain the fairness of the comparison results. However, it is important to note that for specific tasks, we cannot guarantee that methods like OPRO will not exhibit significant performance improvements after exceeding 1600 searches. Given that the primary objective of prompt optimization is to efficiently identify the optimal prompt, we consider a search limit of 300-315 sufficient for evaluating the overall performance of each method.

\paragraph{Model Selection:} In our experiments, we utilized GPT-3.5-turbo and GPT-4 as our task models. While proprietary models like these may undergo upgrades or discontinuation, potentially posing challenges for reproducibility, our results indicate that \textsc{StraGo} performs more effectively with more advanced models. Therefore, we anticipate that \textsc{StraGo} will remain competitive as newer and more sophisticated models become available.

%% file: parts/appendix.tex
\section{More Experiment Details}
\subsection{Data Details} \label{appendix:data}


For the BBH task, we roughly randomly select 50 pieces of data as training data and use the remaining data as the test set. During the experiment, the first 50 pieces of the test set are used as validation data. For natural language understanding tasks and Domain Knowledge tasks, the test set contains more than 1000 pieces of data. We sample 300 or 400 pieces as validation data using stratified sampling. The detailed data division is shown in Table~\ref{tab:split}. For the Industry application task, we choose internal search data. The data part includes: user's location, language, query keywords, some non-personalized query results (including URL, title, and rich text), search history (including query keywords, query time, whether clicked, URL, title, and rich text). The task goal is to measure the extent to which search history supports the reordering of non-personalized queries (1-5), with 1 representing "no help" and 5 representing "Extremely helpful". The detailed data distribution is shown in Figure~\ref{fig:fig3}

\subsection{Prompt Initialization}

In this paper, all prompts are formulated in the Q\_END format, meaning instructions are added after the original question. For example, in the BBH tasks, we append "Let's think step by step" after the question. Table~\ref{tab:prompts} lists the initial prompts for all tasks (for EvoPrompt, we use LLMs to randomly generate several synonymous variants).

\input{tables/data_split}
\input{tables/initial_prompt}

\subsection{Experiment Settings} \label{app:setting}

To fairly compare the performance of different methods, we make appropriate modifications to the official code to ensure that the number of prompt searches for each method is roughly the same. Specifically, for the APO method, we set the candidate set size to 5, with each prompt generating 5 improved versions and 10 synonymous variants per iteration. For OPRO, we set the expansion size to 10. For EvoPrompt, the initial setting is 15 prompts, generating 30 prompts per iteration. To reduce API usage, we use the UCB Bandits algorithm for preliminary screening and retain the top 15 offspring with the highest evaluation scores. This part of the code references the official implementations of \citet{pryzant2023automatic} \footnote{\url{https://github.com/microsoft/LMOps/tree/main/prompt_optimization}}, \citet{yang2023large}\footnote{\url{https://github.com/google-deepmind/opro}} and \citet{guo2023connecting}\footnote{\url{https://github.com/beeevita/EvoPrompt}}. \textsc{StraGo} generates 5 prompts per round and creates 1 synonymous variant for each memory-based prompt. Detailed parameter configurations are provided in Table~\ref{tb:setting}.

\input{tables/setting}
\input{tables/perquery}

\section{Cost Estimate} \label{app:cost}

We refer to the method of \citet{guo2023connecting} for cost estimation. Overall, we divide the entire framework into three stages: generation, filtering, and evaluation. For the estimation of API calls, taking \textsc{StraGo} as an example, the generation stage includes operations such as analyzing experiences, formulating strategies, strategy scoring, strategy optimization, and cross-rewriting. In this stage, each prompt requires about 14 API calls. When using the UCB bandits algorithm for filtering, the number of API calls is $P \times R \times |S|$, where $P$ is the number of prompts generated in each round, $R$ is the number of evaluation rounds, and $|S|$ is the number of samples. In the evaluation stage, each candidate prompt needs to be tested on the validation set, so the total number of calls is $C \times |E|$, where $C$ is the number of candidate prompts retained in each round, and $|E|$ is the size of the validation set.

For the estimation of token consumption, we calculate the average length of each meta prompt, optimized prompt, generated strategy, and task data, and use these values to estimate the total number of tokens $L$ required to generate one prompt in the generation stage. Therefore, the token consumption in the generation stage is $L \times N$, where $N$ is the total search size. In the filtering and evaluation stages, all operations are tested based on task data, so the token consumption of these steps is $(P \times R \times |S|) \times (L_{\text{prompt}} + L_{\text{data}})$. By determining the total number of steps, we can estimate the total token consumption in the filtering and evaluation stages.

\section{Detailed Results of BBH} \label{app:detail_results}
\input{tables/bbh}
Table~\ref{tb:bbh_main_result} presents the overall performance comparison of all methods on GPT-3.5-turbo and GPT-4. This table clearly demonstrates that \textsc{StraGo} outperforms other methods across all tasks. Table~\ref{tb:bbh_analysis} details the modifications each method underwent in 5 BBH tasks, where both \textsc{Acr} and \textsc{Bcr} are calculated based on CoT. The data reveal that in 4/5 tasks, \textsc{StraGo} not only corrected the most errors but also made the fewest incorrect changes. Although \textsc{StraGo}'s performance slightly declines on GPT-3.5-turbo, its highest \textsc{Bcr} or lowest \textsc{Acr} on this model still underscores \textsc{StraGo}'s superior performance.

\section{Meta Prompts} \label{app:meta prompts}
We present all meta prompts used in \textsc{StraGo} in \Cref{tb:prompt1,tb:prompt2,tb:prompt3,tb:prompt4,tb:prompt5,tb:prompt6,tb:prompt7}

\section{Prompt Optimization Results}
We present all prompt optimized by \textsc{StraGo} on Table~\ref{app:res-4} and \ref{app:res-35}.

\input{tables/meta_prompts_v1}

\onecolumn
\scriptsize
\input{tables/prompts}

\normalsize

%% file: tables/data_split.tex
\begin{table}[!ht]
\vspace{-10pt}
\definecolor{Gray}{gray}{0.90}
\newcolumntype{a}{>{\columncolor{Gray}}c}
\centering
\small
\begin{tabular}{@{}l c c c@{}}
\toprule
\textbf{Task}    & \textbf{Train} & \textbf{Eval} & \textbf{Test} \\ \midrule
\textbf{Big-Bench Hard} & & & \\
Logical deduction five objects & 50 & 50 & 200 \\
Movie recommendation & 50 & 50 & 200 \\
Causal judgement & 50 & 50 & 137 \\
Snarks & 50 & 50 & 128 \\
Salient translation error detection & 50 & 50 & 200 \\
\midrule
\textbf{General NLU} & & & \\
SST-5 & 100 & 400 & 1450 \\
TREC & 100 & 300 & 1384 \\
\midrule
\textbf{Domain Knowledge} & & & \\
MEDQA & 100 & 300 & 1173 \\
MEDMCQA & 100 & 400 & 1500 \\
\midrule
\textbf{Industry Application} & & & \\
Personal Query Intent & 50 & 50 & 231 \\
\bottomrule
\end{tabular}

\caption{Data splits.}
\label{tab:split}
\vspace{-10pt}
\end{table}

%% file: tables/initial_prompt.tex
\begin{table}[ht]
\definecolor{Gray}{gray}{0.90}
\newcolumntype{a}{>{\columncolor{Gray}}c}
\centering
\small
\begin{tabular}{@{}cp{0.5\linewidth}}
\toprule
Task    & Initial Prompt  \\ 
\midrule
BBH    & Let's think step by step.  \\ 
\midrule
SST-5    & Select the most accurate emotion description from the options.  \\ 
\midrule
Trec       &    Select the most appropriate type of answer from options.  \\
\midrule
MedQA / MedMCQA       &     Please use your domain knowledge in medical area to solve the questions. \\  
\bottomrule
\end{tabular}

\caption{Initial prompts.}
\label{tab:prompts}
\end{table}

%% file: tables/setting.tex
\begin{table*}[ht]
\centering
\renewcommand\arraystretch{1.6}
\small
\vspace{0.3cm}
\scalebox{0.75}{
\begin{tabular}{|l|c|c|c|c|c|c|c|c|c|c|c|c|}
\hline
\multirow{2}*{\textbf{\makecell[l]{\\Methods}}} &
\multicolumn{4}{c|}{\bfseries Official Search Strategy}& 
\makecell[c]{\bfseries Prompt Updating}&
\multicolumn{5}{c|}{\bfseries  Our Experiments Settings}\\
\cline{2-11}
& {\normalsize \makecell*[c]{Initial \\ size}} 
& \makecell[c]{\normalsize Expansion \\ size per step} 
& \makecell[c]{\normalsize Candidates \\ size per step} 
& \makecell[c]{\normalsize Total \\ Steps} & Method Type & {\normalsize \makecell*[c]{Initial \\ size}} 
& \makecell[c]{\normalsize Expansion \\ size per step} 
& \makecell[c]{\normalsize Candidates \\ size per step} 
& \makecell[c]{\normalsize Total \\ Steps} 
& \makecell[c]{\normalsize Total \\ Search}\\
\hline
\textbf{  APO } & $1$ & $|{P}_{t-1}|\times12$& $4$& $6$ & Explicit Reflection&$1$ & $|{P}_{t-1}|\times15$& $5$&$5$ &$315$ \\

\textbf{  OPRO } & $1$ & $8$ &$-$& $200$&Implicit Reflection& $1$ & $10$ &$-$& $31$ &$310$\\

\textbf{ EvoPrompt } &$10$ & $10$ & $10$ & $10$ & Evolution Algorithm &$15$ & $30$ & $15$ & $10$ &$300$ \\

\textbf{StraGo} & - & - & - & - & Explicit Reflection & 1 & $|{P}_{t-1}|\times10$& $5$&$7$ &$310$ \\

\hline
\end{tabular}}
\caption{Detailed parameter settings.}
\label{tb:setting}
\vspace{-0.2cm}
\end{table*}

%% file: tables/perquery.tex
\begin{figure*}[!t]
    \centering
    \includegraphics[width = \linewidth, trim=0cm 5cm 0cm 0cm, clip=true]{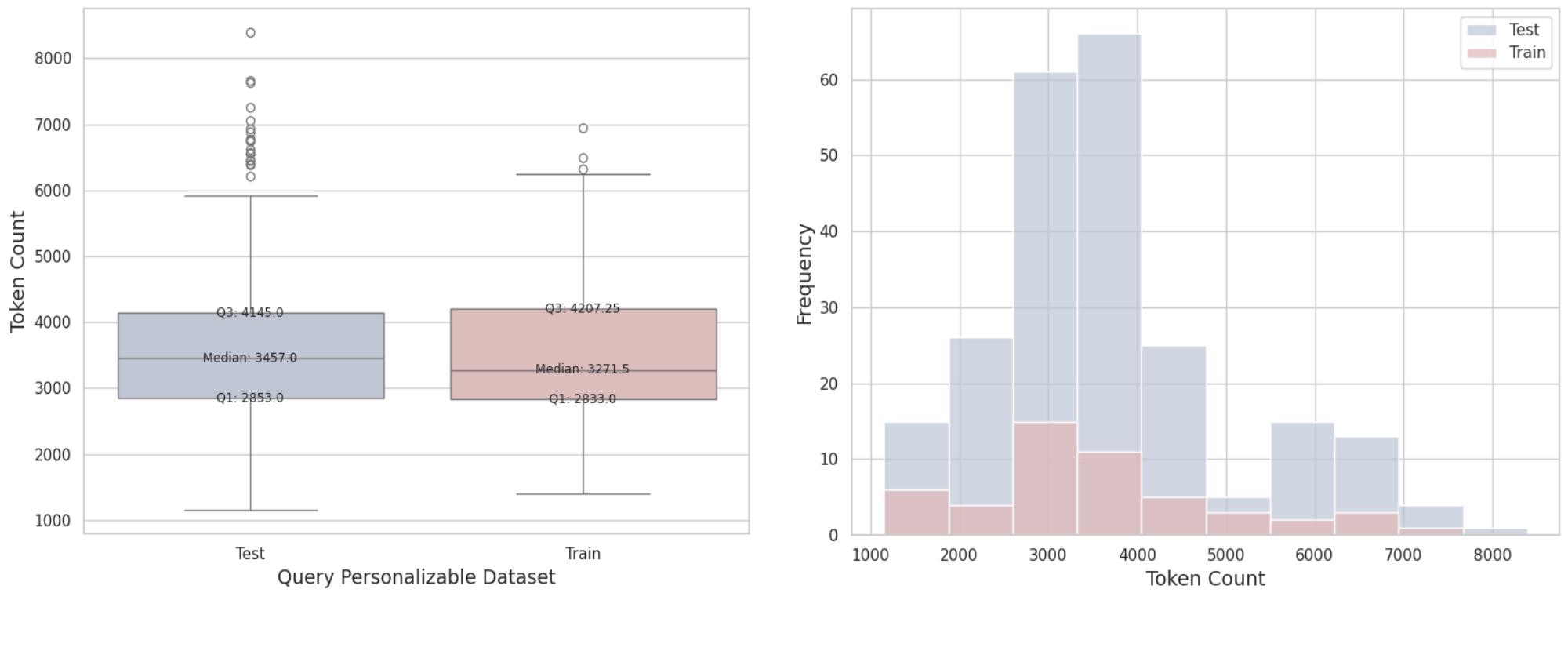}
    \caption{Left side: Distribution of data length in the training and test sets; Right side: Number of data entries in each length interval for the training and test sets.}
    \label{fig:fig3}
\end{figure*}

%% file: tables/bbh.tex
\begin{table*}[!ht]
    \centering
    \setlength\tabcolsep{2pt}
    \scalebox{1.0}{
    \begin{tabular}{cccccccc}
    \toprule
     & \textbf{Method} & \textbf{Logic Dec.} & \textbf{Movie Rec.} &\textbf{ Casual Jud.} & \textbf{Snarks} & \textbf{Salient Trans.} & \textbf{Avg.}\\
    \midrule
    \multirow{5}{*}{\makecell{GPT-3.5\\-turbo}}
    & CoT    & 39.50   & 58.5 & 61.31 & 76.56 & 46.00 & 56.37 \\
    & APO   & 50.50    & 60.00   & 64.96 & 77.34 & 47.00  & 59.96\\
    & OPRO  & 48.5   & 61.50  & 63.50  & 78.90 & 46.50 & 59.78\\
    & EvoPrompt & 47.50 & 61.00 & 64.23 & 78.12 & 47.50 & 59.67\\
    &  \textsc{StraGo} & \textbf{52.00} & \textbf{62.5} & \textbf{66.42} & \textbf{79.69} & \textbf{48.50} & \textbf{61.82} \\
    \midrule
    \multirow{5}{*}{GPT-4}
    &  CoT  & 70.00 & 70.50 & 69.34 & 82.81 & 54.50 & 69.43\\
    & APO  & 81.50 & 78.50 & 70.07  & 85.94 & 66.50 & 76.50\\
    & OPRO & 82.00 & 78.50 & 72.26 & 86.72 & 67.50 & 77.40\\
    & EvoPrompt & 80.00 & 76.00 & 70.80 & 83.59 & 67.00 & 75.48\\
    &  \textsc{StraGo} & \textbf{83.00} & \textbf{81.00} & \textbf{74.45}& \textbf{91.41} & \textbf{69.00} & \textbf{79.77}\\
    \bottomrule
    \end{tabular}
    }
    \caption{The results of GPT-3.5-turbo and GPT-4 on 5 tasks from BBH, with the highest accuracy results highlighted in bold.}
    \label{tb:bbh_main_result}
\end{table*}

\begin{table*}[ht]
    \centering
    \renewcommand{\arraystretch}{1.2}  
    \setlength{\tabcolsep}{5pt}
    \begin{tabular}{lccccccccccc}
        \toprule
        & & \multicolumn{2}{c}{\textbf{Logic Dec.}} & \multicolumn{2}{c}{\textbf{Movie Rec.}} & \multicolumn{2}{c}{\textbf{Casual Jud.}} & \multicolumn{2}{c}{\textbf{Snarks}} & \multicolumn{2}{c}{\textbf{Salient Trans.}} \\
        \cmidrule{3-12}
        \multirow{-2}{*}{\textbf{Model}} & \multirow{-2}{*}{\textbf{Method}} & \textsc{Acr} & \textsc{Bcr} & \textsc{Acr} & \textsc{Bcr} & \textsc{Acr} & \textsc{Bcr} & \textsc{Acr} & \textsc{Bcr} & \textsc{Acr} & \textsc{Bcr} \\
        \midrule
        \multirow{4}{*} {\makecell{GPT-3.5 \\ -turbo}}      & APO       & 32.91 & \textbf{39.67} & 11.97 & 20.48 & 26.19 & \textbf{50.94} & 13.27 & 46.67 & 15.22 & 14.81 \\
        & OPRO      & 35.44 & 37.19 & 12.82 & \textbf{25.30} & \underline{20.24} & 37.74 & 15.31 & \textbf{60.00} & \underline{10.87} & 10.19 \\
        & EvoPrompt & 37.97 & 38.02 & 10.26 & 20.48 & 23.81 & 45.28 & 14.29 & 53.33 & 15.22 & \textbf{15.74} \\
        & \textsc{StraGo}      & \underline{26.58} & 38.02 & \underline{11.11} & \textbf{25.30} & 21.43 & 47.17 & \underline{12.24} & 53.33 & 14.13 & \textbf{16.67} \\
        \midrule
        \multirow{4}{*}{GPT-4}
        & APO       & 12.14 & \textbf{66.67} & 2.84 & 33.90 & 11.58 & 28.57 & 2.83 & 31.82 & \underline{5.50} & 34.07 \\
        & OPRO      & 13.57 & 71.67 &3.55 & 35.59 & 10.53 & 33.33 & 2.83 & 36.36 & 7.34 & 34.07 \\
        & EvoPrompt & 12.86 & 55.88 & 4.96 & 30.51 & 13.68 & \textbf{35.71} & 4.72 & 27.27 & 7.34 & 32.97 \\
        & \textsc{StraGo}      & \underline{10.00} & \textbf{66.67} & \underline{1.42} & \textbf{38.98} & \underline{5.26} & 28.57 & \underline{1.89} & \textbf{59.09} & \underline{5.50} & \textbf{42.86} \\
        \bottomrule
    \end{tabular}
    \caption{Comparative performance of different optimization methods on 5 BBH tasks, measured in terms of \emph{ACR} and \emph{BCR}. A \emph{single underline} denotes the lowest \emph{ACR}, while a \emph{double underline} indicates the highest \emph{BCR}.}
    \label{tb:bbh_analysis}
\end{table*}

%% file: tables/meta_prompts_v1.tex
\begin{table*}[!ht]
    \centering
    \begin{tabular}{@{} >{\raggedright\arraybackslash}p{\linewidth} @{}}
    \toprule
        As a logician, you are good at breaking down the internal logic of the problem step by step. \\
    \\
        <prompt>\{\{prompt\}\}</prompt> \\
     <examples>\{\{examples\}\}</examples> \\
     \\
    I have provided you with a prompt and several examples that include triples of questions, actual answers, and reference answers. Your task is to summarize the \{\{num\}\} most valuable key points to improve your accuracy in solving this type of task.\\
    \bottomrule
    \end{tabular}
    \caption{Collect positive experiences.}
    \label{tb:prompt1}
\end{table*}

\begin{table*}[!ht]
    \centering
    \begin{tabular}{@{} >{\raggedright\arraybackslash}p{\linewidth} @{}}
    \toprule
       As a logician, you are good at breaking down the internal logic of the problem step by step. \\
       \\
    <prompt>\{\{prompt\}\}</prompt> \\
     <examples>\{\{examples\}\}</examples> \\
     \\
    I have provided you with a prompt and several examples that include triples of questions, wrong answers, and reference answers. Your task is to identify \{\{num\}\} primary reasons why the prompt causes these wrong answers.\\
    \bottomrule
    \end{tabular}
    \caption{Collect negative experiences.}
    \label{tb:prompt2}
\end{table*}

\begin{table*}[!ht]
    \centering
    \begin{tabular}{@{} >{\raggedright\arraybackslash}p{\linewidth} @{}}
    \toprule
              As an experienced prompt engineering expert, your task is to evaluate a proposed strategy based on a specific experience. Rate the strategy for its appropriateness, clarity, and effectiveness in addressing the experience. \\
       \\
        \# Experience\\
        <experience>\{\{experience\}\}</experience>\\
        \\
        \# Strategy \\
        <strategy>\{\{strategy\}\}</strategy> \\
        
    \# Rating Criteria \\
    1. Match with Experience(M): The strategy should be directly aimed at mitigating the issue described in the experience. A perfect alignment where the strategy completely addresses the experience issue scores 100 points, whereas a poor match scores lower, depending on how well it addresses the problem. \\
    2. Clarity of Strategy(C): The strategy must be explained clearly and in detail. A strategy that is easy to understand and can be practically implemented by any teacher scores 100 points, while a strategy that is poorly described scores less or 0. \\
    3. Effectiveness in Addressing the Issue(E): Consider how comprehensively the strategy deals with preventing errors and promoting understanding in steps. A strategy that effectively addresses both what should do and what should avoid to minimize errors scores 100 points. A strategy that partially addresses these aspects scores less. \\
    \\
    We asked 5 experts to rate the strategy. Each expert evaluate the strategy independently.\\
    \\
    \# Output Format:
    [\{'M': 78, 'C': 85, 'E': 90\}, \{'M':45,...] \\
    \\
    \# Output
        [\{\\
    \bottomrule
    \end{tabular}
    \caption{Score.}
    \label{tb:prompt3}
\end{table*}

\begin{table*}[!ht]
    \centering
    \begin{tabular}{@{} >{\raggedright\arraybackslash}p{\linewidth} @{}}
    \toprule
        \# Instruction-Score\\
        \{\{instruction\_score\}\} \\
        \\
        Mutate the following instruction reference [\# Instruction-Score] and generate a better instruction.\\
        \\
        \{\{instruction\}\}\\
        \\
        New instruction:\\
    \bottomrule
    \end{tabular}
    \caption{Paraphrase.}
    \label{tb:prompt4}
\end{table*}

\begin{table*}[!ht]
    \centering
    \begin{tabular}{@{} >{\raggedright\arraybackslash}p{\linewidth} @{}}
    \toprule
              As an expert in prompt engineer, your task is to create a step-by-step strategy guide on how to use specific experience based on provided prompt. \\
       \\
        \# Begin Demos \\
        <demo>\\
        <prompt>read the given paragraph and identify the most logical answer among the options.</prompt>\\
        \\
        <example>\\
question: The following paragraphs each describe a set of five objects arranged in a fixed order. The statements are logically consistent within each paragraph. In a golf tournament, there were five golfers: Eve, Amy, Ada, Rob, and Joe. Amy finished second-to-last. Rob finished below Eve. Ada finished above Joe. Joe finished second.\\
Options:\\
(A) Eve finished last\\
(B) Amy finished last\\
(C) Ada finished last\\
(D) Rob finished last\\
(E) Joe finished last\\
Answer: (B) Amy finished last\\
Target: (D) Rob finished last\\
</example>\\
\\
<experience>
One primary reason mistakes occur in this task is due to misunderstanding or misinterpretation of the logical order and relationships presented in the paragraphs
</experience>\\
\\
<strategy>\\
Here is a strategy guide how to achieve "understanding or interpretation of the logical order and relationships":\\
1. Carefully read the entire paragraph to understand the context and the objects or individuals involved.\\
2. Identify the logical relationships or orderings described in the paragraph.\\
3. Create a visual aid such as a list or a diagram. Place the objects or individuals from left to right based on the logical relationships. The leftmost object or individual would be the first in the order and the rightmost would be the last.\\
4. As you read each relationship, adjust the positions of the objects or individuals in your visual aid accordingly.\\
5. Once all relationships have been considered, your visual aid should represent the correct order of the objects or individuals.\\
</strategy>\\
</demo>\\
\\
\{additional demos\}\\
\# End Demos \\ 
     \\
    My current prompt is: \\
    <prompt>\{\{prompt\}\}</prompt> \\
    \\
    And here is the task data:\\
    <example>\{\{example\}\}</example>\\
    \\
    Through comprehensive analysis of the data, I've gained an experience that can improve the prompt:\\
    <experience>\{\{experience\}\}</experience>\\
    \\
    Based on my current prompt, please generate a strategy to address the above experience. \\
    The strategy is: \\
    \bottomrule
    \end{tabular}
    \caption{Generate strategy.}
    \label{tb:prompt5}
\end{table*}

\begin{table*}[!ht]
    \centering
    \begin{tabular}{@{} >{\raggedright\arraybackslash}p{\linewidth} @{}}
    \toprule
             My current instruction is: \\
       <prompt>\{\{prompt\}\}</prompt> \\
       \\
        And Here are some task data: \\
        <example>\{\{example\}\}</example> \\
        \\
        Through comprehensive analysis of the data, I get a experience and corresponding strategy:\\
        \\
        \# Experience \\
        <experience>{{experience}}</experience> \\
        \# Strategy \\
        <strategy>\{\{strategy\}\}</strategy> \\
        \\
        Based on my current prompt, refer to this experience and the strategy, write 1 different improved prompt.
        The improved prompt is:\\
    \bottomrule
    \end{tabular}
    \caption{Optimize.}
    \label{tb:prompt6}
\end{table*}

\begin{table*}[!ht]
    \centering
    \begin{tabular}{@{} >{\raggedright\arraybackslash}p{\linewidth} @{}}
    \toprule
             As an experienced instruction writer, your task is to carefully analyze the provided task cases and instructions in order to generate an improved instruction that will guide an AI system to solve the task more effectively.\\
       \\
       \# Task Cases\\
        The task cases and instructions can be found below: \\
        \{\{examples\}\} \\
        \# Instruction 1 \\
        \{\{prompt1\}\} \\
        \# Instruction 2 \\
        \{\{prompt2\}\} \\
        \\
        Please use the following step-by-step process:\\
        - Step 1: Review the task cases to understand the key objectives and requirements that the instruction needs to address.\\
        - Step 2: Analyze Instruction 1 and identify its strengths and weaknesses in terms of guiding the AI system to solve the task.\\
- Step 3: Perform the same analysis on Instruction 2.\\
- Step 4: Determine how to best combine the strengths of the two instructions while addressing their individual weaknesses.\\
- Step 5: Write an improved, combined instruction that incorporates the insights from the previous steps. The instruction should provide clear guidance for the AI system to solve the task based on the given task cases.\\
- Step 6: Output the improved instruction surrounded by XML tags as follows:\\
<instruction>\\
Your improved instruction goes here.\\
</instruction>\\
    \bottomrule
    \end{tabular}
    \caption{Crossover.}
    \label{tb:prompt7}
\end{table*}

%% file: tables/prompts.tex
\clearpage
\begin{longtable}{l|p{0.7\textwidth}}
\caption{Results on GPT-4.} \label{app:res-4}\\
\toprule
\textbf{Task} & \textbf{Prompt}   \\\midrule
\endfirsthead
\toprule
\textbf{Task} & \textbf{Prompt}   \\\midrule
\endhead
\endfoot
\bottomrule
\endlastfoot 
\nopagebreak
\multirow{5}{*}{logical\_deduction\_five\_objects} & Let\'s approach this problem logically and systematically. Begin by reading the question and the provided statements carefully to understand the order of the objects. Identify the crucial information from each statement, such as the age or position of each object relative to the others. Use this information to construct a preliminary sequence. Then, refine this sequence based on the given relationships. Ensure the final sequence is in line with all the statements. Finally, use this order to answer the question and review your answer to ensure it is logical and consistent with the given information. Let\'s proceed step by step to avoid any mistakes.\\\midrule
\multirow{5}{*}{movie\_recommendation} & Begin by thoroughly understanding the question and the list of movies it provides. Look for common factors like themes, genres, actors, directors, or the time period they were made in. Proceed to meticulously evaluate each option in light of these common elements. Make a well-informed choice and pick the movie that most closely matches these elements.\\\midrule
\multirow{2}{*}{casual\_judgement} & Let\'s dissect the situation in a systematic manner, identifying crucial actions and their subsequent effects, to comprehend the causation and arrive at a logical conclusion.\\\midrule
\multirow{2}{*}{snarks} & Start by meticulously reading the question and all the options to fully comprehend the context. Bear in mind that sarcasm typically involves irony or exaggeration and often presents a situation that contradicts common expectations. Hence, examine the tone and the implicit meaning of each statement. Search for any overstatements or unlikely scenarios in the statements. The statement that seems to convey the opposite of what is logically expected or appears exaggerated is most likely the sarcastic one. Compare all the options, and choose the one that best embodies sarcasm. After selecting, take a moment to reassess your choice to ensure it accurately pinpoints the sarcastic statement. Remember, a good understanding of the subject matter can significantly assist in identifying sarcasm, so take into account the topic context while making your decision.\\\midrule
\multirow{5}{*}{salient\_translation\_error\_detection} & Begin by gaining a thorough understanding of each error category: named entities, numerical values, modifiers or adjectives, negation or antonyms, factual errors, and dropped content. Next, meticulously read the source and translated text, comparing them to identify any discrepancies. Focus on differences in named entities, numerical values, modifiers or adjectives, negation or antonyms, and facts. Based on these discrepancies, identify the type of error and select the corresponding option from the given choices. Remember, the key to accurately identifying the error lies in the details, so be thorough in your examination of each element of the text. After identifying the error, review the content again to ensure no other errors have been missed. If a mistake is found, take the time to understand why it occurred and use this knowledge to avoid similar errors in the future. Repeat this process for each question to ensure consistent accuracy and improvement in task performance.\\\midrule
\multirow{5}{*}{SST-5} & Start by thoroughly examining the given text, keeping a keen eye on the overall tone and context, as well as specific language or expressions that suggest sentiment. Recognize positive sentiment through the presence of words or phrases that indicate approval, satisfaction, or happiness; identify negative sentiment through signs of criticism, unhappiness, or disapproval; and discern neutral sentiment through impartial or unemotional language. Evaluate the intensity of the sentiment, noting that words like 'extremely', 'highly', or 'remarkably' can amplify the sentiment. Be mindful of cultural nuances and language subtleties that could influence the sentiment. After your detailed analysis, select the sentiment description from the provided options that most accurately matches your assessment. Ensure to maintain objectivity and make sure your choice accurately reflects the sentiment present in the text. Once chosen, reassess the text and your selection to confirm there's no misinterpretation or overlooked details.\\\midrule
\multirow{5}{*}{TREC} & For each question presented, your task is to dissect it and identify the type of answer it demands. You should categorize the expected answer into one of these six classifications: (A) Description and abstract concept, (B) Entity, (C) Abbreviation, (D) Human being, (E) Location, or (F) Numeric value. To do this, clarify the main point of the question, focus on its keywords or key phrases, and judiciously examine what kind of response it seeks. Each category signifies a specific nature of response. For clarification, (A) Description and abstract concept refers to explanations, meanings or theories; (B) Entity pertains to organizations, objects or events; (C) Abbreviation denotes acronyms or initials; (D) Human being means names of individuals; (E) Location signifies places; and (F) Numeric value is for numbers, dates or quantities. Your categorization should be based on the type of answer that would optimally satisfy the query. Elaborate on your observation skills and understanding of the question along with the categories to get the correct answer.\\\midrule
\multirow{5}{*}{MedQA} & Firstly, read the question carefully to understand the patient's medical scenario, paying attention to the patient's medical history, symptoms, and test results. Secondly, apply logical analysis to eliminate options that clearly do not align with the given scenario. Thirdly, use your comprehensive understanding of medical terminologies, procedures, and protocols to interpret the medical information provided in the question. Fourthly, employ your medical knowledge and critical thinking to further narrow down the options, considering less obvious connections between symptoms and diseases. Finally, compare the details in the question with the options provided and select the most appropriate answer that represents the best course of action or the most likely diagnosis or treatment based on the comparison. Output the selected answer.\\\midrule
\multirow{5}{*}{MedMcQA} & As a medical professional, your deep comprehension of medical terminologies and principles is paramount for this task. You will come across several multiple-choice questions related to different medical conditions and scenarios. To answer these questions correctly, you need to scrutinize each question and every option carefully, using your medical knowledge to identify and eliminate incorrect options. In cases where you encounter unfamiliar medical terms or situations, rigorous research is advisable. Keep in mind that the devil is in the details and often, medical questions may contain very specific or nuanced details that could be easily overlooked. As part of a continuous learning approach, remember to learn from your past mistakes and apply the lessons to future, unseen medical situations. Regular practice and consistent review of both the mistakes and successful attempts will aid your understanding of medical terminologies and principles drastically. To reduce the scope of errors, take a moment to re-check your answers against any established medical guidelines and principles before submitting them\\
\end{longtable}
\clearpage

\begin{longtable}{l|p{0.7\textwidth}}
\caption{Results on GPT-3.5-turbo.} \label{app:res-35}\\
\toprule
\textbf{Task} & \textbf{Prompt}   \\\midrule
\endfirsthead
\toprule
\textbf{Task} & \textbf{Prompt}   \\\midrule
\endhead
\endfoot
\bottomrule
\endlastfoot 
\multirow{5}{*}{logical\_deduction\_five\_objects} & Begin by thoroughly examining the problem statement, paying attention to the context and the involved parties. Identify any relationships or sequences mentioned and document them. Use a diagram or list to help visualize and structure this data, arranging the parties in the order provided. Once the correct sequence is established, examine each alternative in relation to your determined order or relationships. Select the alternative that best aligns with the given information. Finally, confirm your solution by reviewing your logic and work, making sure no details have been overlooked or mistakes made.\\\midrule
\multirow{5}{*}{movie\_recommendation} & Initiate the process by performing an exhaustive examination of the movies given in the task, identifying shared features like genre, themes, directorial approach, cast, and time of release. Use these noted commonalities as a yardstick for evaluating your choices. If you find any options that you're not well-acquainted with, take a moment to briefly research these movies to grasp their genre, theme, and other vital characteristics. Choose the film that aligns most closely with the shared elements identified in the provided movies. Having a comprehensive knowledge of a vast range of films and the film industry will prove beneficial in this task. Conclude by revisiting your chosen answer to ascertain that it logically corresponds with the detected shared features from the initial list of movies.\\\midrule
\multirow{2}{*}{casual\_judgement} & Begin by thoroughly analyzing the scenario provided in the question, paying close attention to the actions of the individuals involved. Identify the action or event that is being questioned as the potential cause and the subsequent event or result as the effect. Trace the sequence of actions that led to this outcome and evaluate if these actions directly resulted in the outcome. Assess if the identified cause is both necessary and sufficient for the effect to occur. Based on your comprehensive analysis, select the option that accurately answers the question.\\\midrule
\multirow{2}{*}{snarks} & Thoroughly examine the question and the given statements, grasping the context and searching for any inconsistency or contradiction. Assess the tone of each statement, keeping in mind that sarcasm typically exhibits a mocking, ironic, or overstated tone. Contrast the options and utilize these observations to judiciously determine which statement is sarcastic.\\\midrule
\multirow{5}{*}{salient\_translation\_error\_detection} & Begin by gaining a thorough understanding of each error category: named entities, numerical values, modifiers or adjectives, negation or antonyms, factual errors, and dropped content. Next, meticulously read the source and translated text, comparing them to identify any discrepancies. Focus on differences in named entities, numerical values, modifiers or adjectives, negation or antonyms, and facts. Based on these discrepancies, identify the type of error and select the corresponding option from the given choices.\\
\end{longtable}